\documentclass{article}

\usepackage{microtype}
\usepackage{graphicx}
\usepackage{subfigure}
\usepackage{booktabs} 
\usepackage[utf8]{inputenc} 
\usepackage[T1]{fontenc}    
\usepackage{hyperref}       
\usepackage{url}            
\usepackage{booktabs}       
\usepackage{amsfonts}       
\usepackage{nicefrac}       
\usepackage{microtype}      
\usepackage{amsmath,amssymb,mathtools}
\usepackage{gensymb}
\usepackage{xcolor}
\usepackage{bm}
\usepackage{latexsym}
\usepackage{subfigure}
\usepackage{lipsum}
\usepackage{multicol}
\usepackage{color,soul}

\usepackage[accepted]{icmlw2019generalization}

\icmltitlerunning{Kernelized Capsule Networks}

\begin{document}

\twocolumn[
\icmltitle{Kernelized Capsule Networks}



\icmlsetsymbol{equal}{*}

\begin{icmlauthorlist}
\icmlauthor{Taylor Killian}{equal,to}
\icmlauthor{Justin Goodwin}{equal,to}
\icmlauthor{Olivia Brown}{to}
\icmlauthor{Sung-Hyun Son}{to}
\end{icmlauthorlist}

\icmlaffiliation{to}{MIT Lincoln Laboratory, Lexington, MA}

\icmlcorrespondingauthor{Taylor Killian}{taylor.killian@ll.mit.edu}
\icmlcorrespondingauthor{Justin Goodwin}{jgoodwin@ll.mit.edu}

\icmlkeywords{Machine Learning, ICML}

\vskip 0.3in
]



\printAffiliationsAndNotice{\icmlEqualContribution} 

\begin{abstract}
Capsule Networks attempt to represent patterns in images in a way that preserves hierarchical spatial relationships. Additionally, research has demonstrated that these techniques may be robust against adversarial perturbations. We present an improvement to training capsule networks with added robustness via non-parametric kernel methods. The representations learned through the capsule network are used to construct covariance kernels for Gaussian processes (GPs). We demonstrate that this approach achieves comparable prediction performance to Capsule Networks while improving robustness to adversarial perturbations and providing a meaningful measure of uncertainty that may aid in the detection of adversarial inputs.
\end{abstract}

\section{Introduction}
\label{sec:introduction}

Neural networks are a powerful computational paradigm capable of discovering expressive representations of nearly any modality of data. Representation learning with neural networks is provided through optimizing multiple layers of adaptive basis functions. Novel and creative research has introduced highly specialized functions and network architectures applied across a variety of scientific disciplines and commercial industries. However, traditional neural network models, once trained, are deterministic and prone to decaying performance as data distributions shift~\cite{heckman1979sample,sugiyama2007covariate} or adversarial inputs are introduced~\cite{madry2017towards,szegedy2013intriguing}. To overcome these various weaknesses, researchers have developed alternative formulations for neural architectures.

One such neural architecture is the Capsule Network~\cite{hinton2011transforming}. Introduced in 2011, Capsule Networks (CapsNets) replace the scalar ``neurons'' with ``capsules,'' activity vectors that encapsulate the probability of a visual entity being present along with information about the entity's parameters (e.g., pose).  Capsules are expected to provide general viewpoint equivariance across spatial transformations. CapsNets have shown promising results in recent research \cite{hinton2018matrix, kumar2018novel, sabour2017dynamic} on simple image classification problems using a multi-class margin loss. Additionally, these networks have been shown to have some robustness against adversarial perturbations \cite{frosst2018darccc, hinton2018matrix}. Yet, \citet{marchisio2019capsattacks} demonstrate that robustness to adversarial perturbations is not guaranteed with existing architectures.

We aim to improve the robustness of CapsNets by offering an alternative training approach, making use of a Gaussian Process (GP) formed through the construction of a covariance kernel from the capsule output. In addition to the new training approach, the GP classification layer provides a mechanism for detecting adversarial or otherwise ``strange'' inputs to the model through the entropy of the predictive posterior distribution~\cite{bradshaw2017adversarial}. This signal may possibly be utilized in conjunction with the reconstruction approach introduced by~\citet{frosst2018darccc} to identify inputs that the network is uncertain about. Most importantly, we show that a Kernelized Capsule Network (KCN) is more robust to adversarial perturbations while also capable of detecting  examples that lie outside it's learned data distribution.

\section{Background}
\label{sec:background}

\paragraph{Capsule Networks}

\citet{sabour2017dynamic} introduced a dynamic routing by agreement approach to building CapsNets. The hierarchical feature representations are iteratively refined by calculating the agreement between the proposal and input capsules. The activation is simply the length of a given capsule and corresponds to the probability that an entity represented by the capsule is present. The CapsNet parameters are optimized using a margin loss for each output class.  Additionally, a regularization method is applied via an $\ell 2$ reconstruction loss to encourage each capsule to encode parameters of the input image.  In our KCN formulation, we utilize this same reconstruction loss, but provide an alternative to the margin loss through variational inference.

\paragraph{Deep Kernel Learning}

Deep Kernel Learning~\cite{wilson2016stochastic} is an innovative hybrid GP and deep neural network (DNN) approach that utilizes mini-batch training to learn the GP kernel for supervised learning tasks such as image classification. Using a neural network to transform the high dimensional pixel data to a set of local features, $v~=~\mathcal{I}(x)$, a Gaussian Processes model is applied to the set of features via variational learning with inducing points. There are a number of approaches used to solve sparse GP models, e.g., kernel interpolation via KISS-GP~\cite{wilson2015kernel}. Instead of kernel interpolation techniques, we take advantage of the variational learning methods introduced in \cite{titsias2009variational, hensman2015scalable}.  

\paragraph{Adversarial perturbations}

Given a trained network, $\mathcal{I}(\cdot~; \theta)$, an adversarial perturbation to an input, $x$, is a small perturbation, $\eta$, that causes the perturbed input to be misclassified, i.e., $\mathcal{I}(x+\eta; \theta)\neq \mathcal{I}(x; \theta)$. 
Existing methods for computing adversarial perturbations (often referred to as ``adversarial attacks'') range from quick approximations that take only a single gradient step, such as the ``fast gradient sign method'' (FGSM)~\cite{goodfellow2014explaining}, to solving the full optimization as in \citet{carlini2017towards}.  FGSM is useful for evaluating a network's robustness by varying the perturbation size, $\epsilon$, and perturbations are computed as
{\small $\eta = \epsilon~\text{sign}(\nabla_x~\mathcal{L}(x, y))$},
where $\mathcal{L}$ is the loss function used to find adversarial examples.  For white box attacks, we use the loss function for our given network, $L_{\mathcal{I}}(x, y; \theta)$, and for black box attacks, we use the loss function for a surrogate model, $L_{\mathcal{I'}}(x, y; \theta)$.

\section{Kernelized Capsule Networks}
\label{sec:kcn}

We introduce the Kernelized Capsule Network (KCN), a hybrid GP and DNN model that brings together CapsNets and Deep Kernel Learning (DKL). The intent of combining the meaningful representations learned through the CapsNet and the flexible expressivity of GP kernel functions is to more robustly train the CapsNet via the marginal likelihood of the kernel function. Beyond robustness, the KCN has the intrinsic capability of signalling, through the posterior distribution of the induced GP, when the model is uncertain and the predictions it provides are not reliable. 

\begin{figure}[ht]
    \centering
    \includegraphics[width=.5\textwidth]{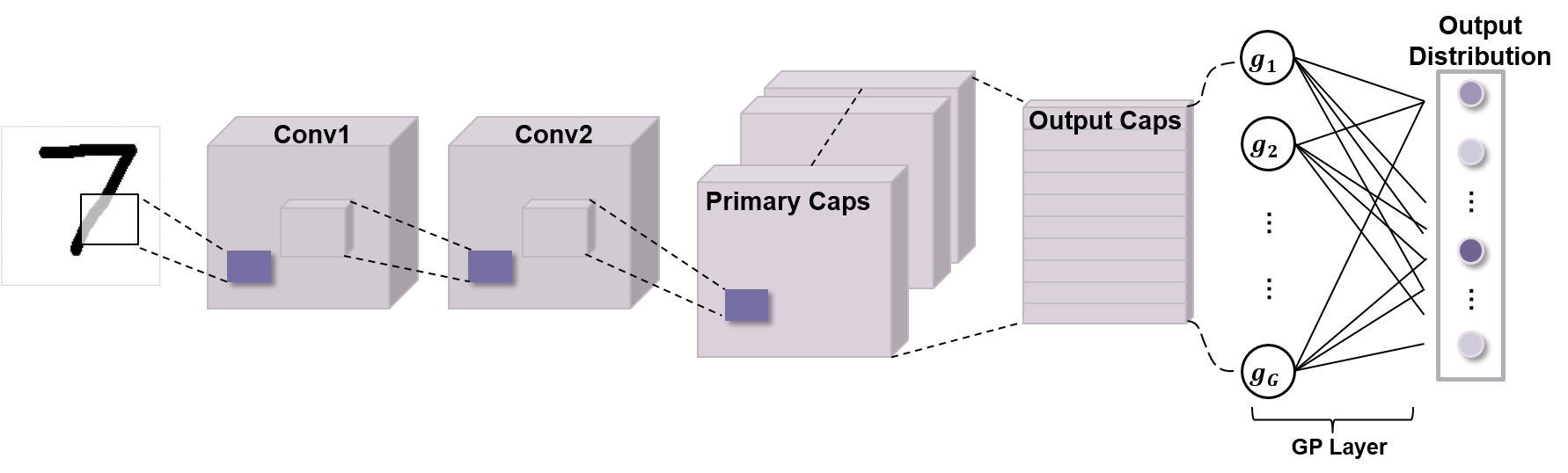}
    \caption{The Kernelized Capsule Network}
    \label{fig:kcn_arch}
\end{figure}

A KCN (Figure~\ref{fig:kcn_arch}) is built from the nominal CapsNet architecture~\cite{sabour2017dynamic}, with two convolutional input layers at the head of the primary capsule layer. Intermediate feature representations from the primary capsules are converted into output capsules, {$\mathcal{V} = \{v_1,...v_{N_c}|v_j \in\mathbb{R}^{k \times 1}\}$}, through the dynamic routing procedure. We treat the set of output capsules, {\small $\mathcal{I}(x) = \mathcal{V}$}, as a final feature representation that is used to construct a GP kernel function (Section~\ref{sec:var_learn}), where the correlation of individual features are measured. This induced GP is then used to infer the probability of the input image's label by way of a Softmax likelihood (eqn.~\ref{eqn:gp_mll}).

\subsection{Variational Learning}
\label{sec:var_learn}

The challenge to learning with capsule networks is providing a meaningful loss when given a vector instead of the standard logits of neural networks.  Given capsules $v_j\in\mathcal{V}$, we want to construct a loss that helps capture the learning goal of capsule networks (equivariance to spatial transformation) while accounting for the correlations between features and providing robustness to adversarial perturbations.  To achieve this goal we utilize Gaussian Processes, namely Deep Kernel Learning.  The input data will be the set of capsules $\mathcal{V}$ with the goal of estimating the class probabilities.  

To address the scalability of GPs, we use stochastic variational inference to approximate the matrix inversion and factorization operations used in calculating the predictive posterior distribution~\cite{titsias2009variational, wilson2015kernel}.  First, define the kernel function $K_{xz}=k(X, Z)$ where,
{\small
\begin{equation}\label{eqn:rbf}
k(X, Z) = \mathrm{exp} (-\gamma \| \mathcal{I}(X) - \mathcal{I}(Z) \|^2),
\end{equation}}
and $\gamma$ is a free parameter. Then, as shown in~\citet{hensman2015scalable}, the approximate posterior Gaussian Process is defined by $q(f) = \mathcal{N}(f| \mu, \Sigma)$ where the mean is
{\small
\begin{equation}\label{eqn:approximate_gp_mean}
\mu = k(X, Z) k(Z, Z)^{-1}m = K_{xz}K^{-1}_{zz} m
\end{equation}}
and the covariance is
{\small
\begin{equation}\label{eqn:approximate_gp_cov}
\Sigma = K_{xx} - K_{xz}(K^{-1}_{zz} - K_{zz} S K^{-1}_{zz})K_{zx} )
\end{equation}}
where the $m$ and $S$ are the variational parameters of the distribution $q(u)=\mathcal{N}(u |m, S)$, while $Z$ represents the learned inducing variables. The parameters are optimized by maximizing the bound on the marginal log likelihood for $N$ data points,
\vspace{-.25cm}
{\small
\begin{equation}\label{eqn:gp_mll}
log(p(y)) \geq \sum_{n=1}^N \mathbb{E}_{q(f_n)}[\log(p(y_n|f_n)] - KL[q(u)\|p(u)]
\end{equation}}
This is a summation across the marginal probabilities because we can factor the likelihood for classification as
{\small
\begin{equation}\label{eqn:gp_factored_lhood}
p({\bf y} | f, X) = \prod_n p(y_n | f(x_n))
\end{equation}}
For our classification problem we utilize the Softmax likelihood using the categorical distribution.

\subsection{Model Usage}

With a fully trained KCN, images are first input to the CapsNet.  The capsule representations, $\mathcal{V}$, are then passed through the kernel, which via eqt.~\ref{eqn:approximate_gp_mean} and eqt.~\ref{eqn:approximate_gp_cov} construct a set of correlated GPs. The latent functions produced by these GPs are then passed through the marginal log likelihood (eqt.~\ref{eqn:gp_mll}) to estimate the classification probabilities. Finally, predictions are made by taking the $argmax$ of that output distribution, with the expectation that the highest probability corresponds to the correct class of the input image.

\subsection{Detection of Adversarial Inputs}

\citet{frosst2018darccc} demonstrated that an auxiliary decoder network, $d(I(x))$, could be trained to reconstruct the input image from the capsule features corresponding to the true class $\{v_j~|~j = y\}$ (see Figure~\ref{fig:decoder} in the appendix). This enables the detection of adversarially perturbed inputs, as perturbations corrupt the features of the capsule representation, affecting the $\ell 2$ accuracy of the reconstruction when compared to the original input. \citet{bradshaw2017adversarial} showed that an analogous mechanism could be developed for a standard deep neural network (DNN) by measuring the entropy of the posterior predictive distribution trained through a hybrid GP-DNN. Motivated by these two papers, we seek to investigate how each signal can contribute to the detection of adversarial inputs. 

To detect adversarial inputs, we assume ``normal'' network behavior when the reconstruction error is less than or equal a given threshold and perturbed images are detected when the threshold is exceeded. We examine Receiver Operating Characteristic (ROC) curves to understand the detection performance.  Similarly, we investigate using a threshold on the entropy of the network's predictions to detect adversarial inputs. A reported entropy value above the threshold signifies the network's uncertainty when faced with inputs that it is unfamiliar with or have been otherwise corrupted.


\section{Experiments}
\label{sec:experiments}

Our primary experimental objective is to demonstrate the robustness of the KCN in comparison to the CapsNet when faced with both white and black box FGSM attacks with varying $\epsilon \in [0,1]$. To isolate the effect that the non-parametric GP layer has on the performance of the KCN, we also compare our final architecture (CapsNet foundation~$+$~GP classification layer~$+$~decoder network) to a version of KCN that forgoes the auxiliary reconstruction network, featuring only the GP classification layer. We refer to this model as KCN-GP.

The model used to develop black box FGSM attacks is a simple two-layer CNN. Specifics about each model architecture and the training/testing protocols are included in the appendix, Section~\ref{apdx:train}.

\subsection{Datasets}

We evaluate each model described above against MNIST~\cite{lecun2010mnist}, SVHN~\cite{netzer2011reading}, and CIFAR10~\cite{krizhevsky2009learning}. Details about train/test splits, training augmentation, and other preprocessing are included in Section~\ref{apdx:train}. All reported results are from performance on the test set.

\subsection{Detecting Adversarial Perturbations}

\begin{figure}[H]
    \centering
    \subfigure[MNIST-$\ell 2$]{\label{fig:mnist_l2}  \includegraphics[height=2.5cm]{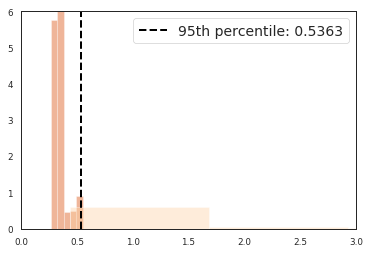}}
    \hspace{-0.15cm}
    \subfigure[MNIST-Entropy]{\label{fig:mnist_ent}  \includegraphics[height=2.5cm]{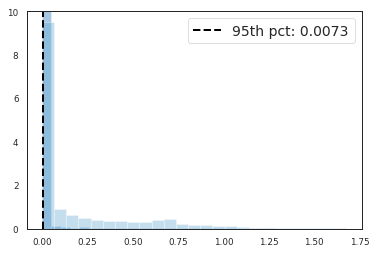}}
\caption{ MNIST test set detection signals. Darker shading corresponds to unperturbed test examples while the lighter shading corresponds to examples perturbed with FGSM ($\epsilon=0.3$). }
\label{fig:mnist_hist}
\end{figure}

Figure~\ref{fig:mnist_hist} shows the nominal $\ell 2$ distances and entropies of a trained KCN model before and after perturbation (dark and light shading, respectively; $\epsilon=0.3$) for the MNIST test set. The dashed vertical line indicates a notional threshold drawn to allow a 5\% false alarm rate.  Histograms for other datasets and values of epsilon are included in Section~\ref{apdx:hist}.

Immediately, one can recognize that the perturbed distribution of $\ell 2$ is separated from the unperturbed values. Note that the threshold drawn for the entropy values is near zero, leaving a long tail of unperturbed inputs that will almost always mix with the values of perturbed inputs 

These two observations preliminarily suggest that entropy may not be as helpful of an adversarial detection signal compared with the $\ell 2$ reconstruction error. We directly compare the efficacy of each detection method through ROC curve analysis in the next section.

\begin{figure*}[ht]
\centering
\hspace{-.25cm}
\subfigure[MNIST-Accuracy]{\label{fig:mnist_accuracy}  \includegraphics[height=3.5cm]{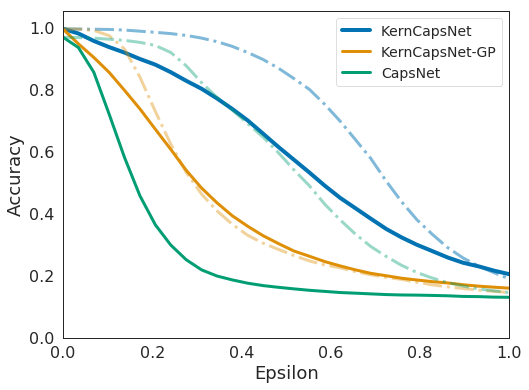}}
\hspace{-0.15cm}
\subfigure[SVHN-Accuracy]{\label{fig:svhn_accuracy}  \includegraphics[height=3.5cm]{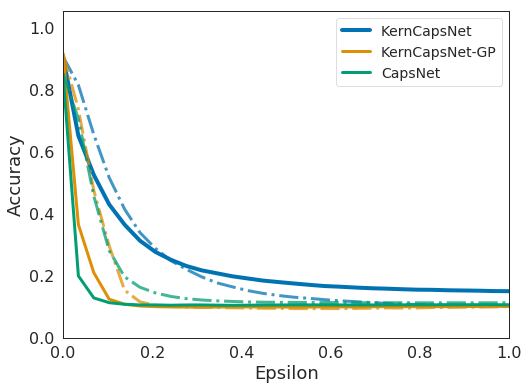}}
\hspace{-0.15cm}
\subfigure[CIFAR10-Accuracy]{\label{fig:cifar_accuracy}  \includegraphics[height=3.5cm]{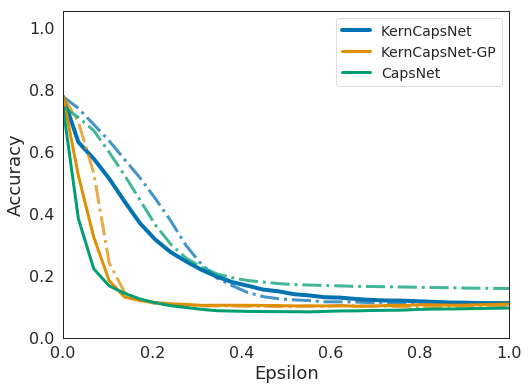}}\\
\subfigure[MNIST-ROC]{\label{fig:mnist_roc}  \includegraphics[height=3.5cm]{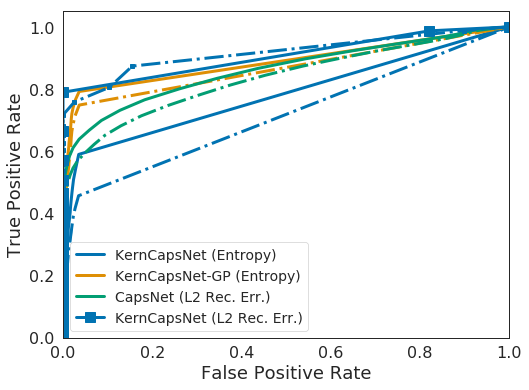}}
\hspace{-0.15cm}
\subfigure[SVHN-ROC]{\label{fig:svhn_roc}  \includegraphics[height=3.5cm]{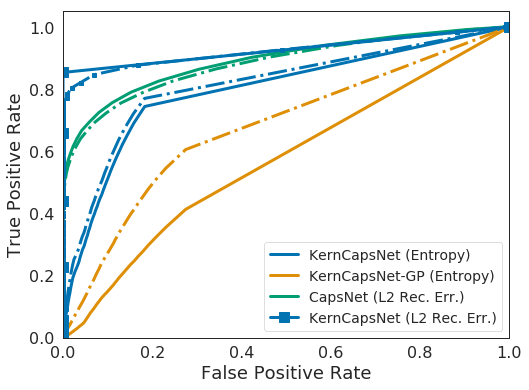}}
\hspace{-0.15cm}
\subfigure[CIFAR10-ROC]{\label{fig:cifar_roc}  \includegraphics[height=3.5cm]{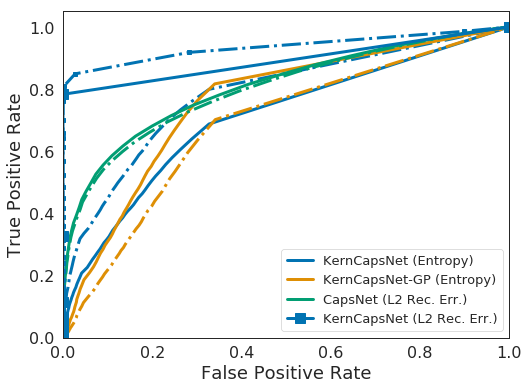}}\\
\caption{ (a)-(c):Accuracy of the models when facing adversarial perturbations of varying magnitude. (d)-(f):ROC curves for each attack detection signal. (across all images, the dashed line of the same color corresponds to a black box attack on that model)}
\label{fig:results}
\end{figure*}

\section{Results}
\label{sec:results}

The results of our experiments are represented across Figure~\ref{fig:results}, as well as in Tables~\ref{tab:results} and~\ref{tab:auc} in Section~\ref{apdx:kcn_results}. 
For each dataset, the accuracy of the KCN is very close to--and often greater than--that of the CapsNet (refer to Table~\ref{tab:results}), showing that using a marginal log likelihood as a loss function can effectively train a CapsNet with auxiliary structures (in this case, the GP classifier and decoder network) without sacrificing performance. This indicates that the induced kernel function provides a useful mechanism to flexibly transform the vector quanitity set of capsule outputs to a Softmax likelihood, rather than relying on hand-tuned loss functions such as the Hinge or Margin losses used in prior work~\cite{sabour2017dynamic,frosst2018darccc}.

\begin{figure}[H]
    \centering
    \includegraphics[width=0.465\textwidth]{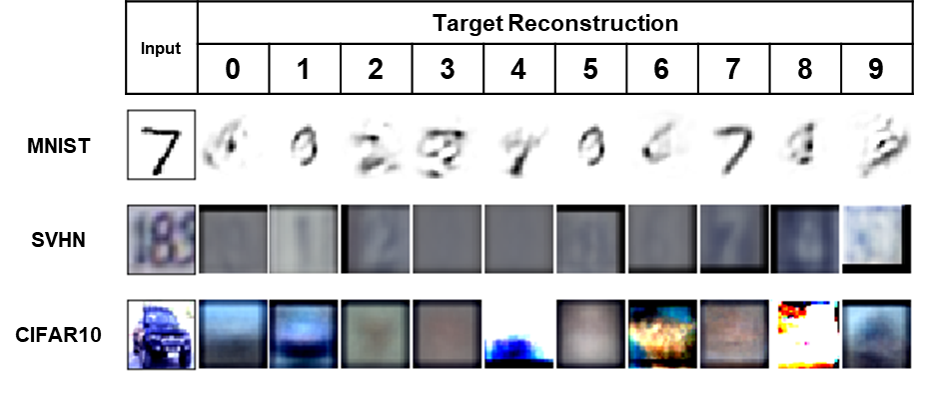}
    \caption{Sample reconstructions from each dataset. MNIST capsules appear to look for "strokes", SVHN and CIFAR10, given limited feature resolution, provide more detail on color and color gradients. Capsules do appear to have "attention" behavior, e.g., the SVHN example, only the 8 is reconstructed.}
    \label{fig:samples}
\end{figure}

Figure~\ref{fig:results}(a)-(c) demonstrates that the KCN is more robust than the nominal CapsNet against both white and black box FGSM attacks (solid and dashed lines, respectively) across the three datasets presented in this paper. It appears that the $\ell 2$ reconstruction error serves as an effective regularizer, improving robustness against adversarial perturbations, particularly against black-box attacks (see Figure~\ref{fig:samples} for example reconstructions made by the KCN). 

When evaluating which detection signal is most salient to identifying adversarial perturbations, Figure~\ref{fig:results}(d)-(f) indicates that the $\ell 2$ reconstruction distance is generally effective distinguishing between perturbed and nominal inputs. AUC metrics for each approach are reported in Table~\ref{tab:auc}. Of note, the $\ell 2$ detection signal from the KCN is stronger than that of the CapsNet, evidenced by the higher reported AUCs. 

Although the KCN, by virtue of the GP output layer, does signal that the classifier is uncertain in the presence of adversarially perturbed inputs, entropy alone is not sufficient to determine when an attack occurs due to the overlap of the calculated entropy distributions between normal and perturbed data. The entropy signal could possibly serve as an early warning of model degradation where the $\ell 2$ metric would provide a more dependable signal of whether or not an input to the model has been corrupted.

\section{Conclusion}
\label{sec:conclusion}

This paper introduces the Kernelized Capsule Network, a model that leverages the flexibility of non-parametric kernel functions to extend the rich representations learned through a CapsNet. We've shown that this approach to training a CapsNet maintains performance, while improving model robustness as well as the ability to detect adversarially perturbed inputs. We are intrigued by the relative simplicity afforded to the training of CapsNets through the GP marginal log likelihood.  We believe that hybrid models in the vein of the KCN, or DKL more generally, provide an exciting avenue for future research in model robustness, interpretability and compositionality.

\section*{Acknowledgements}
Foremost, the authors would like to thank the MIT LL ICE Line committee for their support in funding this work. We are grateful to Peter Morales and Austin Jones for many helpful conversations and suggestions as well as Dan O'Connor and Anu Myne for their review and support of this work.

\bibliography{paper}
\bibliographystyle{icml2019}

\appendix
\section{Training details}
\label{apdx:train}

Our models are built using PyTorch~\cite{paszke2017automatic}. Deep Graph Library ~\cite{wang2018dgl} is used to solve the dynamic routing problem for Capsule Networks.  The probabilistic framework Pyro~\cite{bingham2018pyro} is used for Deep Kernel Learning. For training we utilize the standard train and test sets for each dataset.  We transform the datasets by random shifting images by up to 4 pixels and then apply normalization.  The normalizing is calculated using the standard values of mean and standard deviation for each dataset: MNIST has $\mu= 0.1307$ and $\sigma=0.3081$, CIFAR10 has $\mu = (0.5071, 0.4867, 0.4408)$ and $\sigma=(0.2675, 0.2565, 0.2761)$, and SVHN has $\mu= (0.5, 0.5, 0.5)$ and $\sigma=(0.5, 0.5, 0.5)$.  The architecture and training protocol are outlined below.

\subsection{Architecture}
\label{apdx:proc_arch}
The capsule network, $\mathcal{I}(x)$, consists of two simple convolutional layers that are unique to each dataset. For MNIST we use $\text{C}(1, 12, 4, 2, 3)$ - $\text{C}(12, 16, 3, 2, 1)$ where $\text{C}(n, k, s, p)$ is a convolution layer followed by a rectified linear unit (ReLU) non-linearity for all but the last layer with parameters of $n$ output channels, kernels of size $k \times k$, stride $s$, and padding $p$. The output feature map of these convolutional layers are passed to the Primary Capsule Layer where there are $8$ convolutional capsules, $\{\text{C}_i\text{(16, 32, 8, 2, 0)}\}_{i=1}^8$. Each capsule is then stacked and flatten so that there are $8$ capsule vectors followed by the squashing non-linearity defined in~\citet{sabour2017dynamic}. Next we compute the output capsules using the dynamic routing by agreement algorithm implemented in the Deep Graph Library~\cite{wang2018dgl}.  This produces the output capsules $v_j \in \mathbb{R}^k$ where $k=16$ for each class $j=1 \dots  N_c$ where $N_c=10$ for each of the datasets. For SVHN and CFIAR10 the convolutional layers are C(3, 64, 4, 2, 1) - C(64, 64, 3, 2, 1) with convolutional capsule layers defined by $\{\text{C}_i\text{(64, 32, 8, 2, 0)}\}_{i=1}^8$. See Figure~\ref{fig:kcn_arch}.

\begin{figure}[H]
    \centering
    \includegraphics[width=0.25\textwidth]{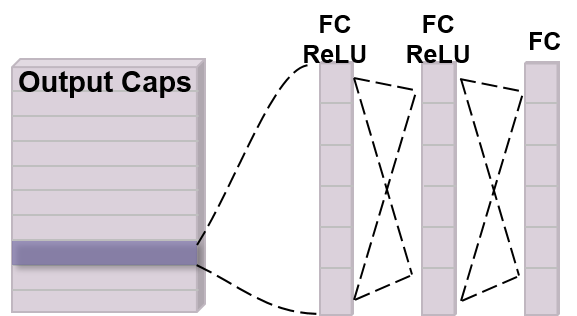}
    \caption{Decoder structure used to reconstruct the input from the Output Capsule layer representation.}
    \label{fig:decoder}
\end{figure}

The reconstruction layer, see Figure~\ref{fig:decoder} is a set of linear layers of sizes $\text{L}(k N_c, 512)$-$\text{L}(512, 1024)$-$\text{L}(1024, i_c i_h i_w)$ where $i_{c,h,w}$ are the number of image channels and the image height and width respectively. The first two linear layers are followed by the ReLU non-linearity.

The GP classification layer utilizes the RBF kernel (Equation~\ref{eqn:rbf}) that expects a vector dimension of $k N_c$, the total number of elements across all capsules. The number of inducing variables is set to $70$ and the output size of the variational distribution $q(u)$ is set to the number of class $N_c$.

\subsection{Training Protocol}

For each minibatch Pyro~\cite{bingham2018pyro} provides the ELBO loss, defined by the function ''Trace Mean Field ELBO'', for deep kernel learning.  In addition to the ELBO loss, we add the reconstruction loss using a multiplicative factor of $100$ due to the different scales between the losses. The Adam~\cite{kingma2014adam} optimizer is used with a learning rate of $1\mathrm{e}{-2}$ and default PyTorch parameters.

\section{Additional KCN Performance Comparisons}
\label{apdx:kcn_results}

See Table~\ref{tab:results} and Table~\ref{tab:auc} below for performance comparisons.

\begin{table*}[ht]
\caption{{\bf Summary of Training Results}}
\label{tab:results}
\resizebox{\textwidth}{!}{%
\begin{tabular}{@{}lccccccccccc@{}}
\cmidrule(r){2-12}
                        & \multicolumn{3}{c}{MNIST} &  & \multicolumn{3}{c}{SVHN} &  & \multicolumn{3}{c}{CIFAR10} \\ \cmidrule(lr){2-4} \cmidrule(lr){6-8} \cmidrule(l){10-12} 
    & Accuracy  & Avg. Ent. & Avg. $\ell 2$  &  & Accuracy  & Avg. Ent. & Avg. $\ell 2$ &  & Accuracy      & Avg. Ent.    & Avg. $\ell 2$  \\ \cmidrule(lr){2-4} \cmidrule(lr){6-8} \cmidrule(l){10-12} 
CapsNet         & 96.7 &  --   &  .349  &  & 85.0 &   --   &  .114  &  & 74.3  &  --  & .455     \\
KCN-GP          & 99.3 & .0062  &    --  &  & 91.4 &  .059  &    --  &  & 77.7  & .068 &   --     \\
KCN             & 99.4 & .0067  &  .339  &  & 90.6 &  .033  &  .220  &  & 77.6  & .061 & .977     \\ \cmidrule(l){2-12} 
\end{tabular}}
\end{table*}

\begin{table*}[ht]
\caption{{\bf Summary of Perturbation Detection AUC}}
\label{tab:auc}
\resizebox{\textwidth}{!}{%
\begin{tabular}{@{}lccccccccccc@{}}
\cmidrule(r){2-9}
                        & \multicolumn{2}{c}{MNIST} &  & \multicolumn{2}{c}{SVHN} &  & \multicolumn{2}{c}{CIFAR10} \\ \cmidrule(lr){2-3} \cmidrule(lr){5-6} \cmidrule(l){8-9} 
                        & White Box  & Black Box & & White Box  & Black Box &  & White Box & Black Box  \\ \cmidrule(lr){2-3} \cmidrule(lr){5-6} \cmidrule(l){8-9} 
CapsNet ($\ell 2$)  &  0.8616 & 0.8344   &  & 0.8907 &  0.8813   &  & 0.8073  &  0.7992   \\
KCN  ($\ell 2$)     & {\bf 0.9072} & {\bf 0.9160}  &  & {\bf 0.9266} &  {\bf 0.9229}  &  & {\bf 0.8915} & {\bf 0.9350}      \\
KCN (Entropy)       & 0.7806 & 0.7134  &  & 0.7953 &  0.8132  &  & 0.7062  & 0.7843      \\
KCN-GP (Entropy)    & 0.8860 & 0.8631  &  & 0.5688 &  0.6758  &  & 0.7580  & 0.6844      \\
\cmidrule(l){2-9} 
\end{tabular}}
\end{table*}

\section{Histograms}
\label{apdx:hist}

\begin{figure}[H]
    \centering
    \subfigure[SVHN-$\ell 2$]{\label{fig:svhn_l2}  \includegraphics[height=2.5cm]{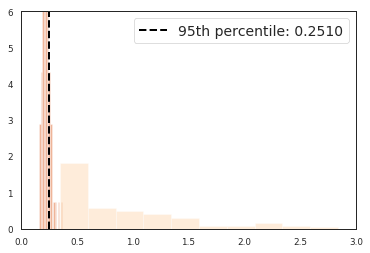}}
    \hspace{-0.15cm}
    \subfigure[SVHN-Entropy]{\label{fig:svhn_ent}  \includegraphics[height=2.5cm]{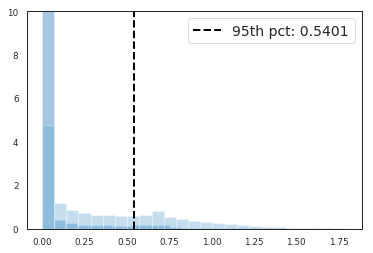}}
\caption{ SVHN test set detection signals.  Darker shading corresponds to unperturbed test examples while the lighter shading corresponds to examples perturbed with FGSM ($\epsilon=0.3$) }
\label{fig:SVHN_hist}
\end{figure}

\begin{figure}[H]
    \centering
    \subfigure[CIFAR10-$\ell 2$]{\label{fig:cifar_l2}  \includegraphics[height=2.5cm]{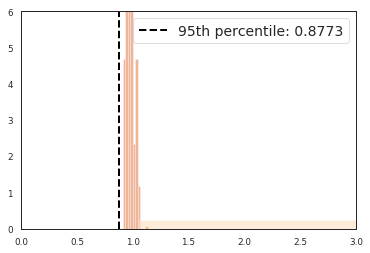}}
    \hspace{-0.15cm}
    \subfigure[CIFAR10-Entropy]{\label{fig:cifar_ent}  \includegraphics[height=2.5cm]{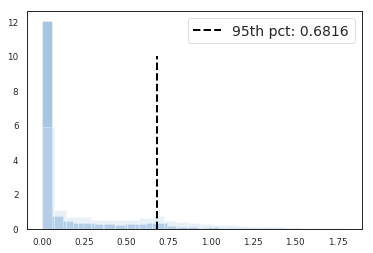}}
\caption{ CIFAR10 test set detection signals.  Darker shading corresponds to unperturbed test examples while the lighter shading corresponds to examples perturbed with FGSM ($\epsilon=0.3$). }
\label{fig:CIFAR_hist}
\end{figure}

\begin{figure*}
    \centering
    \includegraphics[width=\textwidth,height=\textheight]{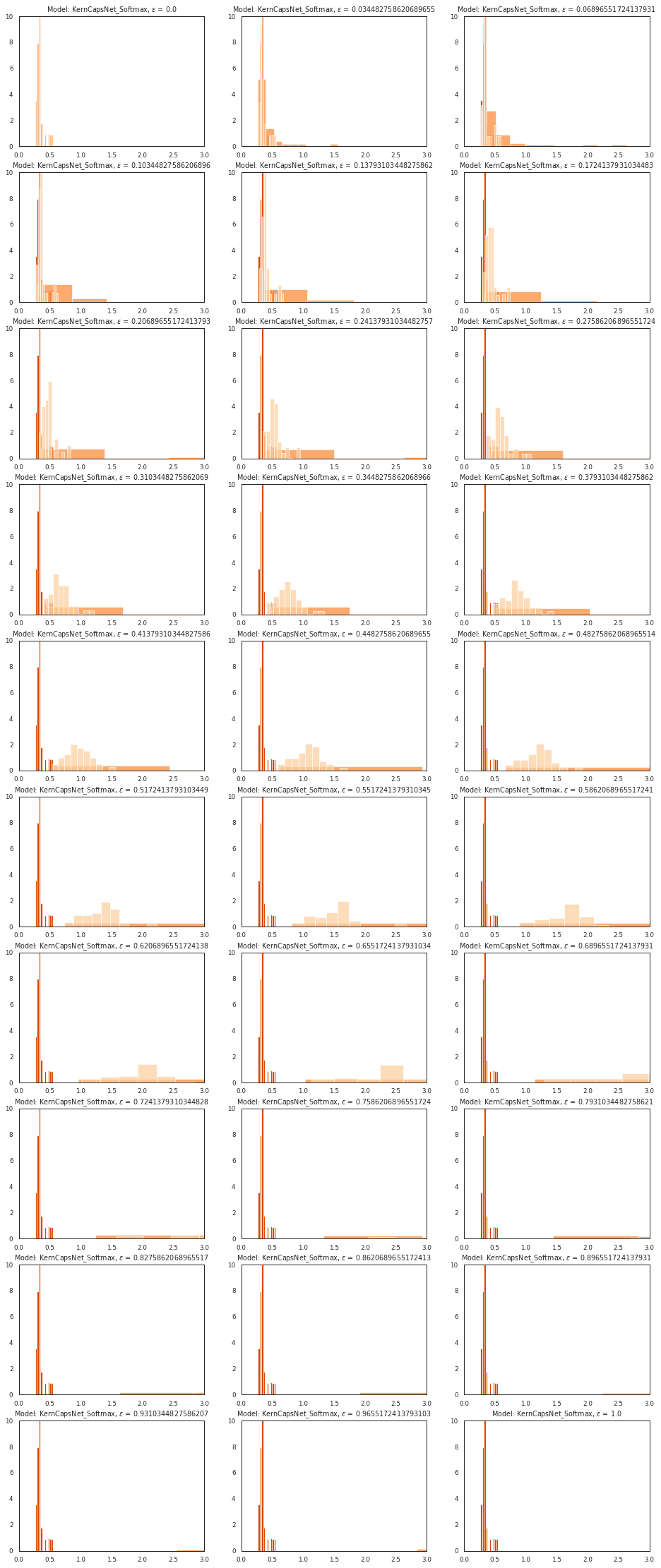}
    \caption{MNIST Compilation of $\ell 2$ Reconstruction Error Histograms for all values of $\epsilon$. The two lighter shades correspond to both white and black box attacks.}
    \label{fig:mnist_full_l2_hist}
\end{figure*}

\begin{figure*}
    \centering
    \includegraphics[width=\textwidth,height=\textheight]{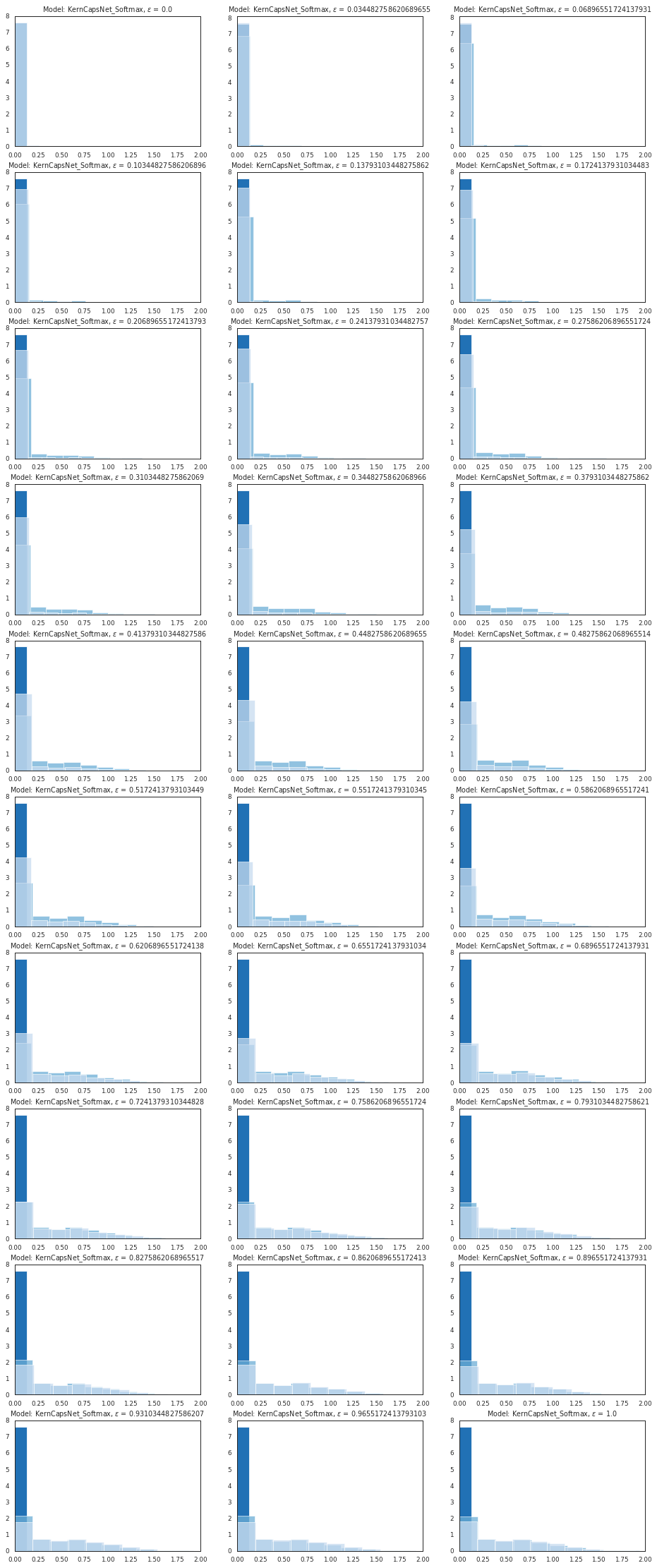}
    \caption{MNIST Compilation of Entropy Histograms for all values of $\epsilon$. The two lighter shades correspond to both white and black box attacks.}
    \label{fig:mnist_full_ent_hist}
\end{figure*}

We present additional histograms of the $\ell 2$ reconstruction error and entropy values that can be used to distinguish perturbed inputs from nominal, unperturbed inputs. In Figure~\ref{fig:SVHN_hist} and Figure~\ref{fig:CIFAR_hist} we present the histograms of the normal, unperturbed inputs and inputs perturbed by FGSM with $\epsilon=0.3$ for SVHN and CIFAR10 respectively. Figure~\ref{fig:mnist_full_l2_hist} and Figure~\ref{fig:mnist_full_ent_hist} show, for MNIST when using a KCN model, how the distribution of $\ell 2$ reconstruciton errors and Entropy values change as $\epsilon$ increases. From these images, it is easier to gain an intuition into how effective a chosen threshold can be in detecting adversarially perturbed inputs. For $\ell 2$ the distributions separate quite early and completely, making it possible for an effective choice in threshold to be made. This task becomes much harder when considering the entropy values as there is significant overlap of the heavy tails of the normal, unperturbed distribution. This complicates any choice of threshold as there will likely be several nominal inputs that will be incorrectly classified as adversarial.

\newpage

\end{document}